\documentclass[a4paper, 10pt, conference]{IEEEtran}
\ifCLASSINFOpdf
  \usepackage[pdftex]{graphicx}
  
\else
\fi
\usepackage{algorithmic}
\usepackage{algorithm}
\usepackage{xcolor}
\usepackage{booktabs}
\usepackage{longtable}
\usepackage{makecell}

\begin{document}
%
\title{Explaining Neural Network Model for Regression}

\author{\IEEEauthorblockN{Mégane Millan}
\IEEEauthorblockA{
Sorbonne Universite, CNRS UMR 7222, ISIR\\
F-75005, Paris, France\\
Email: millan@isir.upmc.fr}
\and
\IEEEauthorblockN{Catherine Achard}
\IEEEauthorblockA{
Sorbonne Universite, CNRS UMR 7222, ISIR\\
F-75005, Paris, France\\
Email: catherine.achard@sorbonne-universite.fr}
}


%


\maketitle

\begin{abstract}
Several methods have been proposed to explain Deep Neural Network (DNN). 
However, to our knowledge, only classification networks have been studied to try to determine which input dimensions motivated classification decision. 
Furthermore, as there is no ground truth to this problem, results are only assessed qualitatively in regards to what would be meaningful for a human.

In this work, we design an experimental database where ground truth is reachable: we generate ideal signals and disrupted signals with errors and train a neural network that determines the quality of said signals.
This quality is simply a score based on the distance between the disrupted signals and the corresponding ideal signal.
We then try to find out how the network estimated this score and hope to find time-steps and dimensions of the signal where errors occurred. 
This experimental setting enables us to compare several methods for network explanation and to propose a new method, named Accurate GRAdient (AGRA), based on several trainings, that decreases noise present in most state-of-the-art results.
Comparative results show that the proposed method outperforms state-of-the-art methods for locating time-steps where errors occur in the signal.
\end{abstract}


%
\IEEEpeerreviewmaketitle

\section{Introduction}
Machine learning is increasingly present in today's life since the arrival of the first Convolutionnal Neural Networks (CNN) \cite{726791}. 
Performances achieved by such networks are impressive and have led to their development in many applications, such as smart vehicles. 
Despite these performances, errors still exist and can have dramatic consequences, especially for applications where lives are at stake.
Furthermore, in medical fields, for example, it is desirable not only to have a final classification result but also to know the causes of the decision. 
For all these reasons, more and more research is being conducted on DNN explanation, as mentioned in recent literature reviews \cite{dovsilovic2018explainable}, \cite{samek2017explainable}, \cite{zhang2018visual}. 

To our knowledge, all these methods try to explain DNN trained for classification task: the goal is to find out which elements of the input led to the decision of the network. 
Unfortunately, no ground truth exists.
Therefore, network-explanation results are only evaluated by looking at the produced maps and comparing them to what a human operator believes to be correct. 
Without an objective tool that quantifies results, it is difficult to compare the results of different methods.

In this article, we propose to build an experimental setup, associated with a ground truth, to quantify explanation results of networks.
This setup aims at estimating signals quality: we created a database of ideal signals to which errors were added at random positions.
A note is associated to each signal, depending on the distance of an example to its ideal version.
A CNN is trained in regression to find this note. 
Then, the network explanation aims at determining which part of the input (temporal position and dimension) occasioned the score provided by the network. 
Such a setup with a ground truth enables us to compare quantitatively different DNN-interpretability algorithms.

In order to determine time-steps and dimensions of the input signal where errors occurred, we do a gradient descent that transforms the input to a signal that has the best possible note.
This gradient descent enables us to have a gradient according to the input signal. 
Such a strategy is not new and it is known that these gradients are very noisy \cite{smilkov2017smoothgrad}, \cite{kim2019saliency}. 
During our experiments, we have found that these gradients vary a lot depending on training and weight initialisation.
Actually, training the same model several times on the same database leads, for a given input example, to gradients that change a lot from one model to another. 
Some model gradients find some errors but not others, and some are very noisy while others are not, etc.
We chose to take advantage of these variations to estimate an "accurate gradient" from all the models. 
The proposed method, named Accurate GRAdient (AGRA), consists in averaging the gradients generated by the different trainings and weight initialization for the same input signal.

Thanks to our experimental setup, we quantitatively compare AGRA with several gradients-based methods and show its efficiency.
Moreover, AGRA can be combined to other gradient-based methods to improve their performance.

Thus, two main contributions are proposed in this article.
First, we develop an experimental database that allows to qualitatively and quantitatively compare DNN explanation methods. 
Second, we introduce a new DNN explanation technique AGRA, based on gradients, that outperforms state-of-the-art methods.

\section{Related work}
\subsection{Explaining Deep Neural Networks methods}
Several methods exist in the literature to explain DNN. 
Their goal is to find the contribution of each input feature to the output and thus, to produce attribution maps.
Methods can be grouped into three main categories: class activation based approaches, perturbation based approaches and gradient based approaches.

\subsubsection{Class activation based approaches}
Methods such as Class Activation Map (CAM) \cite{zhou2016learning}, Gradient-weighted Class Activation Mapping (Grad-CAM) \cite{selvaraju2017grad}, or Uncertainty based Class Activation Maps (U-CAM) \cite{patro2019u} propose to generate Class Activation Maps that highlight pixels of the image the model used to make the classification decision.
The goal is thus to produce maps similar to human attention regions.
These maps are estimated in a multi-class classification context and are class-discriminative. 

\subsubsection{Perturbation based approaches}
The idea of these approaches is to disturb some portions of the input image and look at their influence on the output.
Work in \cite{zeiler2014visualizing} consists in systematically occulting different portions  of  the input  image  with  a  grey  square,  and  monitoring  the output  of the  classifier. 
As  the  probability  of  the  correct  class  drops  significantly when the object is occluded, this technique localizes objects in  the  scene.
Another approach, based on perturbation, proposed by Ribeiro \textit{et al.} \cite{ribeiro2016should}, is the Local Interpretable Model-Agnostic Explanation (LIME).
A model is explained by perturbing the input and constructing a local linear model that can be interpreted. 
Thus, LIME makes local  approximations  of  the  complex decision surface.

\subsubsection{Gradient based approaches}
Simonyan et al. \cite{simonyan2013deep} proposed to compute sensitivity maps as the gradient of the output according to input pixels in a classification task.
If $S_c(x)$ is the score function of the classification network for the class $c$ and input image $x$, then sensitivity maps are defined as:
\begin{equation}
    M_c(x) = \frac{\partial S_c(x)}{\partial x}.
\end{equation}
By intuition, important gradient values correspond to locations in the image that have a strong influence on the output. 

In practice, these sensitivity maps are very noisy.
A first solution to improve them is to change the back-propagation algorithm.
Thus, deconvolution networks \cite{zeiler2014visualizing} and Guided Backpropagation \cite{zeiler2014visualizing} propose to discard negative gradient values during the back-propagation step.
The idea is to keep only entries that will have a positive influence on the score.

Another problem with gradient-based techniques is that the score function $S_c$ may saturate for important input characteristics \cite{sundararajan2016gradients}.
Thus, the function may be flat (but important) around these inputs and thus, has a small gradient.
Some methods address this problem by computing the global importance of each pixel. 
Thus, DeepLIFT (Deep  Learning  Important FeaTures)   \cite{shrikumar2017learning} decomposes   the output prediction  by  back-propagating   contributions  of  all  neurons  in  the  network  to  every feature of   the   input. 

Layer-wise relevance propagation (LRP)  \cite{bach2015pixel} uses a pixel-wise decomposition to understand the contribution of each single pixel  of the input image $x$ to the score function $S_c(x)$. 
A propagation rule, applied  from the output back to the input, distributes class relevance found at a given layer onto the previous layer.
It leads to a heatmap that highlights pixels responsible for the predicted class.

Three other methods, based on the classical back-propagation algorithm, exist to explain DNN: Gradient $\times$ Input \cite{shrikumar2017learning}, Integrated gradient \cite{sundararajan2017axiomatic} and SmoothGrad \cite{smilkov2017smoothgrad}. 

Gradient $\times$ Input \cite{shrikumar2017learning}, \cite{bach2015pixel} was proposed to improve attribution maps. 
They are simply computed as the product between the gradients of the output with respect to the input and the input itself:
\begin{equation}
\label{eq:GradInput}
    GradInput(x) = M_c(x) \times x
\end{equation}

Instead of computing the gradients of the output according to the input pixels $x$, Sundararajan et al. \cite{sundararajan2017axiomatic} integrate the gradients along a path from a baseline $x'$ to the input $x$. The integrated gradient, for the $i^{th}$ dimension of the input $x$ is defined as:
\begin{equation}
\label{eq:intgrad}
    IntGrad_i(x) = (x_i-x'_i)\times \int_{1}^{\alpha=0} \frac{ \partial{ S_c(x' + \alpha (x-x'))} }{\partial x_i} d\alpha
\end{equation}
where $\frac{\partial{ S_c(x)} }{\partial x_i}$ is the gradient of $S_c$ according to $x$ along the $i^{th}$ dimension.

During computation, the integral is approximated via a summation:  gradients at the N points lying on the straight line from the baseline $x'$ to the input $x$, are added.
Integrated gradients add up to the difference between the outputs $S_c$ at $x$ and the baseline $x'$. 
Thus, if the baseline has a near-zero score, integrated gradients form an attribution map of the prediction output $S_c(x)$.

Given the rapid fluctuations of the gradient for an input image $x$, it is less meaningful than a local average of gradient values. 
Thus, SmoothGrad \cite{smilkov2017smoothgrad} proposes to create an improved sensitivity maps based on a smoothing of $\partial{ S_c(x)}$ with a Gaussian kernel.
As the direct computation of such a local average in a high-dimensional input space is intractable, Smilkov et al. compute a stochastic approximation by taking random samples in a neighborhood of the input $x$ and averaging the resulting sensitivity maps:
\begin{equation}
\label{eq:SmoothGrad}
    SmoothGrad(x) = \frac{1}{N} \sum \limits_{i=0}^N M_c(x + \mathcal{N}(0,\sigma^2))
\end{equation}
where N is the number of noised inputs, and $\mathcal{N}(0,\sigma^2)$ is a Gaussian noise with a $0$ mean and a $\sigma$ standard deviation.

In this article, we also propose to use a gradient-based approach and to denoise the so-obtained gradient. 
The proposed approach, based on several trainings, can be combined to other gradient-based methods to improve their performances.

\section{AGRA method to obtain accurate gradient}
In this work, we first propose to design an experimental setup to explain DNN. 
Then, we introduce a new method allowing to denoise the gradient of the output according to an input using several trainings of the same DNN. 

\subsection{Designing an experimental setup}
A problem often encountered with DNN explanation algorithms is the lack of ground truth. 
It is therefore difficult to quantitatively estimate the performance of such algorithms. 
To address this issue, we design a setup where this ground truth is available.
This setup is composed of 2D temporal signals.
Both dimensions are generated using sinusoids with different lengths, to which a small Gaussian noise has been added. 
These signals represent ideal signals in the database. 
Then we artificially create perturbations in both dimensions by adding high-frequency Gaussians. 
The number of perturbations varies uniformly between 0 and 8 and their position and the dimension where they appear are also drawn according to a uniform law. 

A score, re-scaled between $0$ and $10$, is then given to each of these signals.
This score is based on the Mean Square Error (MSE) between the signal without perturbation and the disrupted signal. 
$0$ is attributed to ideal signals while score gets close to $10$, when many perturbations are present.
1000 signals are thus generated, 750 are used for training and 250 for testing, drawn according to a uniform law.

\begin{figure}[h]
    \includegraphics[width=\linewidth]{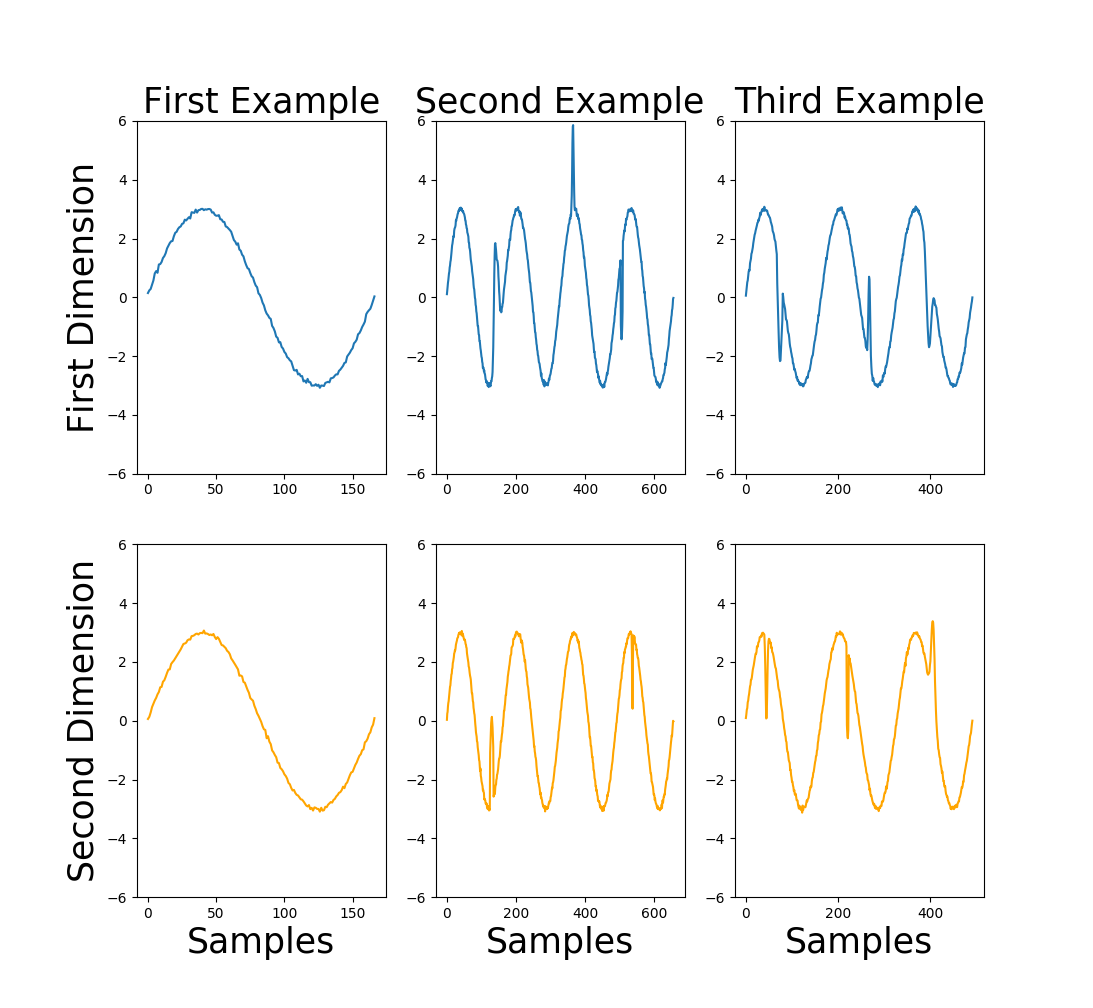}
    \caption{Examples of one ideal signal (left) and two perturbed signals (middle with 5 pertubations and right with 6 pertubations)}
    \label{fig:signals}
\end{figure}

The goal of the network will then be to regress the score of each input signal while the goal of the DNN explanation will be to find time-steps and dimensions of the errors.
Three examples of signals extracted from the database are presented in Figure~\ref{fig:signals}.

Even if we are working on synthetic examples with a ground truth regarding the DNN explanation, this setup corresponds to a real application that aims to determine the quality of gestures in sports \cite{MILLAN2020} or surgical context \cite{fawaz2018evaluating}, for instance.
In addition to assigning a score, DNN explanation will make it possible to determine where gestures are poorly carried out.

\subsection{The AGRA method}
First a CNN is trained to regress the scores with a MSE loss between the predicted scores $\hat{s}(x)$ and the scores of the ground truth $s(x)$: $l_1(x)=(\hat{s}(x) - s(x))^2$.
Then, for DNN explanation, a gradient of the output according to the input example $x$, as that proposed in \cite{simonyan2013deep}, is computed, without changing the weights of the networks.
It is used to change the input $x$ so that its note increases.
As the goal is to find differences difference between ideal signals and perturbed ones, the loss used for gradient back-propagation is the MSE between the predicted score and the optimal note ($0$ in our case): $l_2(x)=(\hat{s}(x) - 0)^2$. 
Several iterations are done until the ideal note is reached as explained in Algorithm~\ref{alg1} where $\lambda$ is the learning rate and $\epsilon$ is the tolerance: loop stops when the loss is below $\epsilon$.

\begin{algorithm}
\caption{Compute Gradient}
\label{alg1}
\begin{algorithmic}
\REQUIRE $x, \lambda, \epsilon$
\ENSURE $GRAD(x)$
\STATE $x' = x$
\WHILE{$l_2(x') >\epsilon $} 
    \STATE $grad = \frac{ \partial {l_2(x')}}{\partial x'}$
    \STATE $x' \leftarrow x' - \lambda \times grad$
\ENDWHILE
\STATE GRAD(x) = x - x'
\end{algorithmic}
\end{algorithm}

\begin{figure*}
    \centering
    \includegraphics[width=0.7\textwidth]{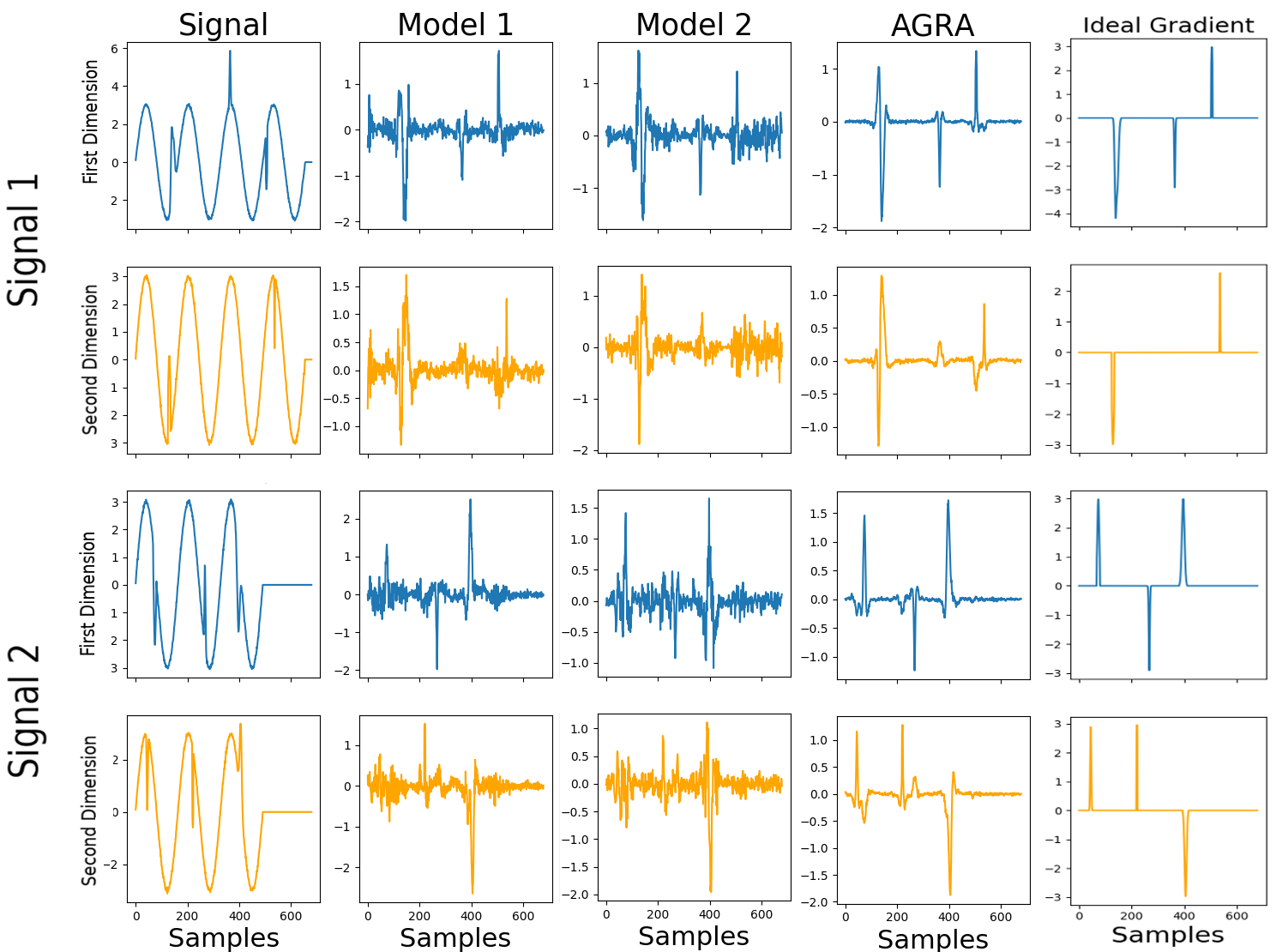}
    \caption{Gradients obtained from two  models with different initialization and with the AGRA method.}
    \label{fig:moy_grads}
\end{figure*}

Unfortunately, and as stated before, this gradient is very noisy \cite{smilkov2017smoothgrad}, \cite{kim2019saliency}. 
Moreover, during our experiments, we observed that it depends significantly on weights initialisation and training of the network.
Thus, even if two different trainings lead to similar regression scores, gradients are highly variable. 
Two examples of gradient can be found on Figure~\ref{fig:moy_grads}.

We decided to take advantage of these variations and average  gradients of different models with different trainings, to obtain a noise-reduced and more accurate gradient.
So, we trained $N$ times the same network to obtain $N$ models. 
Let $GRAD_i$, the gradient of the output according to the input, obtained with model $i$, as described in the algorithm~\ref{alg1}.
AGRA is then obtained as described in Algorithm~\ref{alg2}. 
AGRA method needs several trainings of the same model, which is computationally expensive.
However, as shown in Figure~\ref{fig:moy_grads}, the so-obtained gradients are more accurate.
Moreover, they no longer depend on training and initialisation, which was the case before when either good or bad gradients were obtained.

\begin{algorithm}
\caption{Compute Accurate Gradient: AGRA}
    \label{alg2}
    \begin{algorithmic}
    \REQUIRE $x$
    \ENSURE $AGRA(x)$
    \STATE $grad = 0$
    \FOR {$it = 0$ to $N$} 
        \STATE $grad = grad + GRAD_{it}(x)$
    \ENDFOR
    \STATE $AGRA(x) = grad /N$
    \end{algorithmic}
\end{algorithm}

\section{Experimental results}
For all methods involved in this section, we use the loss function $l_2(x)$ previously defined to compute gradients. 

\subsection{Training procedure}
The regression network consists of four temporal convolutional layers with $(8, 8, 16, 16)$ filters of size $(25, 5, 5, 5)$, with no bias added.
Each of them is followed by a pooling layer with size $(3, 3, 3, 3)$. 
Two fully connected layers with $50$ and $1$ neurons end the neural network with between them a dropout layer with a $0.5$ probability, with no bias.
The network is learnt with adam optimizer \cite{Kingma2015AdamAM} and a $0.01$ learning rate, for $100$ epochs. 
The network regresses a score between $0$ and $10$ and is trained $50$ times to obtain $50$ models.
The mean MSE across the $50$ models, on the test set, is of 0.619 with a standard deviation of 0.089. 
So, during prediction, these models have a similar behavior.


\subsection{Qualitative results}
Firstly, we present qualitative results of the five following methods:
\begin{itemize}
    \item Classical gradient GRAD \cite{simonyan2013deep} computed with Algorithm~\ref{alg1}, a learning rate of 0.1 and a tolerance $\epsilon$ of 0.015.
    \item GRAD $\times$ input $x$ as defined in equation~\ref{eq:GradInput} and proposed by \cite{shrikumar2017learning}, \cite{bach2015pixel}
    \item Smooth gradient \cite{smilkov2017smoothgrad} estimated as the mean of $50$ gradients  obtained with Algorithm~\ref{alg1} by adding a Gaussian noise with $0$ mean and $0.1$ as standard deviation on the input signal (equation~\ref{eq:SmoothGrad}).
    \item Integrated gradient \cite{sundararajan2017axiomatic}. As the proposed network has no bias, the baseline $x'$ is fixed to a zero signal with the same length than $x$. In these conditions, the score of the baseline is $s(x')=0$ and integrated gradients can been interpreted as an attribution map of the prediction output $s(x)$. Integrated gradients have already been multiplied by the input as explained in equation~\ref{eq:intgrad}.
    \item The AGRA method with $50$ trained models.
\end{itemize}

\begin{figure*}
  \includegraphics[width=.95\textwidth]{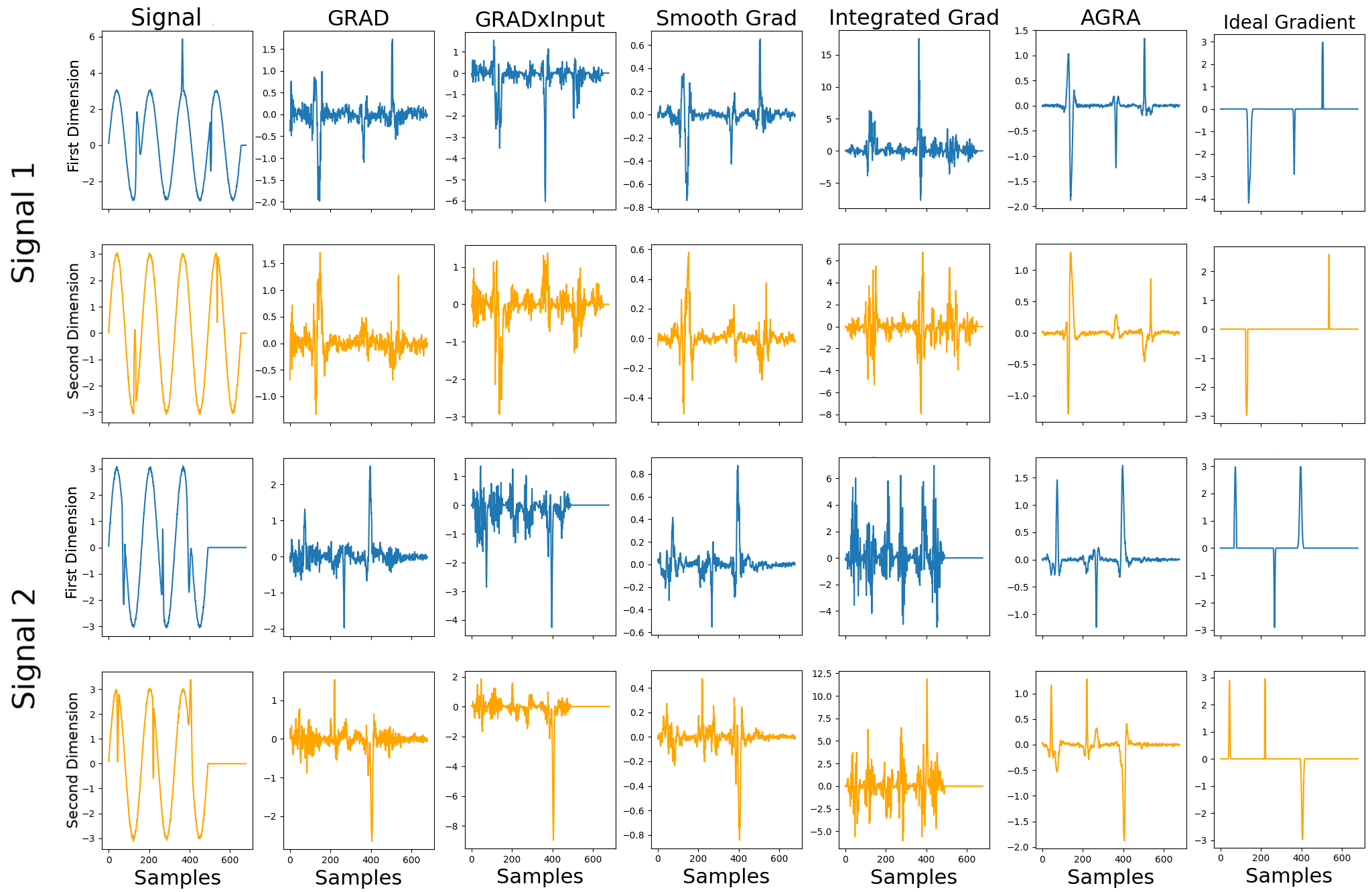}
  \caption{Results obtained with all the methods on the same two examples.}
  \label{fig:all_res}
\end{figure*}

As shown in Figure~\ref{fig:all_res}, classical gradients (GRAD) are noisy and do not lead to clear and easy to interpret results, since peaks at perturbation locations are sometimes too thin and small and can be considered as noise.
Furthermore, multiplying these noisy gradients with the input only makes the results worse.
Indeed, interesting peaks are enhanced but global results appear noisier than before.
Moreover, the sign of the gradient, which gives information on the direction of the error, is lost due to this multiplication.
Using smooth gradient instead of classical gradient gives better qualitative results with considerably less noise than before.
However, noise is still present and the results are again difficult to interpret. Moreover, the magnitude of the gradient is often under-estimated.
Integrated gradients are very noisy and have peaks at undisturbed positions, making them very difficult to interpret. As they are multiplied by the input signal, the sign of the gradient is lost.
As shown in Figure~\ref{fig:all_res}, less noisy and more accurate results are achieved with AGRA method. 
Gradients actually highlight the locations corresponding to perturbations and have the correct direction to reconstruct the ideal signal.


\subsection{Quantitative results}
To compare methods more thoroughly, giving quantitative results is crucial.
Since ground truth is available for each example, it is possible to compute ideal gradients (the difference between perturbed signals and ideal ones) and compare them with results obtained with the different methods. 
Two metrics are used to make this comparison: 
\begin{itemize}
    \item Mean Squared Error (MSE) between the signal without errors and the reconstructed signal obtained thanks to the gradients. This metric cannot be used for methods such as GRAD$\times$Input or Integrated Gradient, since their goal is only to highlight important time-steps and not to reconstruct a perfect signal.
    \item Pearson correlation coefficient between the ideal gradient and the gradient obtained with the different methods. To avoid penalising methods, that do not manage the signs (GRAD$\times$Input and Integrated Gradient), this coefficient is computed between the norms of both ideal gradient and gradient from the methods. 
\end{itemize}

The $250$ training examples have been averaged to obtain these metrics. Moreover, for GRAD, GRAD$\times$Input, Smooth Grad and Integrated Gradient, metrics have been computed on the 50 trained models and afterwards averaged. 

\begin{table}[]
    \centering
    \begin{tabular}{c|c}
        \hline
         \bfseries Methods & \bfseries Mean Squared Error \\
         \hline
         GRAD \cite{simonyan2013deep} & 7.65 \\
         Smooth Grad \cite{smilkov2017smoothgrad} & 7.85 \\
         AGRA & \bfseries 5.06 \\
         \hline 
    \end{tabular}
    \vspace{.3cm}
    \caption{Mean Squared Error between ground truth gradients and estimated ones for different methods. 
    }
    \label{tab:mse}
\end{table}
Table~\ref{tab:mse} presents the MSE obtained with different methods. As a reminder,  an estimated gradient fitting perfectly to the ground truth one would correspond to a 0 MSE.
Both GRAD and Smooth Grad methods are noisy.
Moreover, Smooth Grad does not keep gradient magnitude.
Thus, AGRA method outperforms both of these methods according to MSE. 
AGRA is therefore the most suitable method for signal reconstruction. 

\begin{table}[]
    \centering
    \begin{tabular}{c|c}
        \hline
         \bfseries Methods & \bfseries Pearson Correlation \\
         \hline
         GRAD \cite{simonyan2013deep} & 0.81 \\
         GRADxInput \cite{shrikumar2017learning}, \cite{bach2015pixel} & 0.82 \\
         Smooth Grad \cite{smilkov2017smoothgrad} & 0.79  \\
         Integrated Gradient \cite{sundararajan2017axiomatic} & 0.55 \\
         AGRA & \bfseries 0.94 \\
         \hline
    \end{tabular}
    \vspace{.3cm}
    \caption{Pearson Correlation coefficient for different methods estimated between the norm of gradients}
    \label{tab:corr}
\end{table}
As shown in Table~\ref{tab:corr}, Pearson correlation coefficients vary between 0.55 and 0.94.
As Pearson correlation coefficients are standardised (the correlation is divided by the standard deviation of both gradients), they can be estimated in a meaningful way for each method, even when the gradient is multiplied by the input.
The best results are obtained with our proposed method, which confirms the previous qualitative study and proves that this method gives better results than other state-of-the-art methods.

Table~\ref{tab:corr1} gives the Pearson coefficients obtained by keeping the sign of the gradients when calculating the correlation: the correlation is estimated for each of the two dimensions and then averaged. 
Using this metric, only Grad and Smooth Grad methods can be evaluated since for the other two, multiplying by the input will change signs of gradient and results will not be exploitable.
AGRA is again the most efficient method, even if Pearson coefficient do not take into account gradient magnitude, which does not penalize Smooth Grad as the MSE did.

\begin{table}[]
    \centering
    \begin{tabular}{c|c}
        \hline
         \bfseries Methods & \bfseries Pearson Correlation \\
         \hline
         GRAD \cite{simonyan2013deep} & 0.68 \\
         Smooth Grad \cite{smilkov2017smoothgrad} & 0.66  \\
         AGRA & \bfseries 0.84 \\
         \hline
    \end{tabular}
    \vspace{.3cm}
    \caption{Pearson Correlation coefficient for different methods estimated on all gradient dimension}
    \label{tab:corr1}
\end{table}

\begin{figure}[h]
    \centering
    \includegraphics[width=\linewidth]{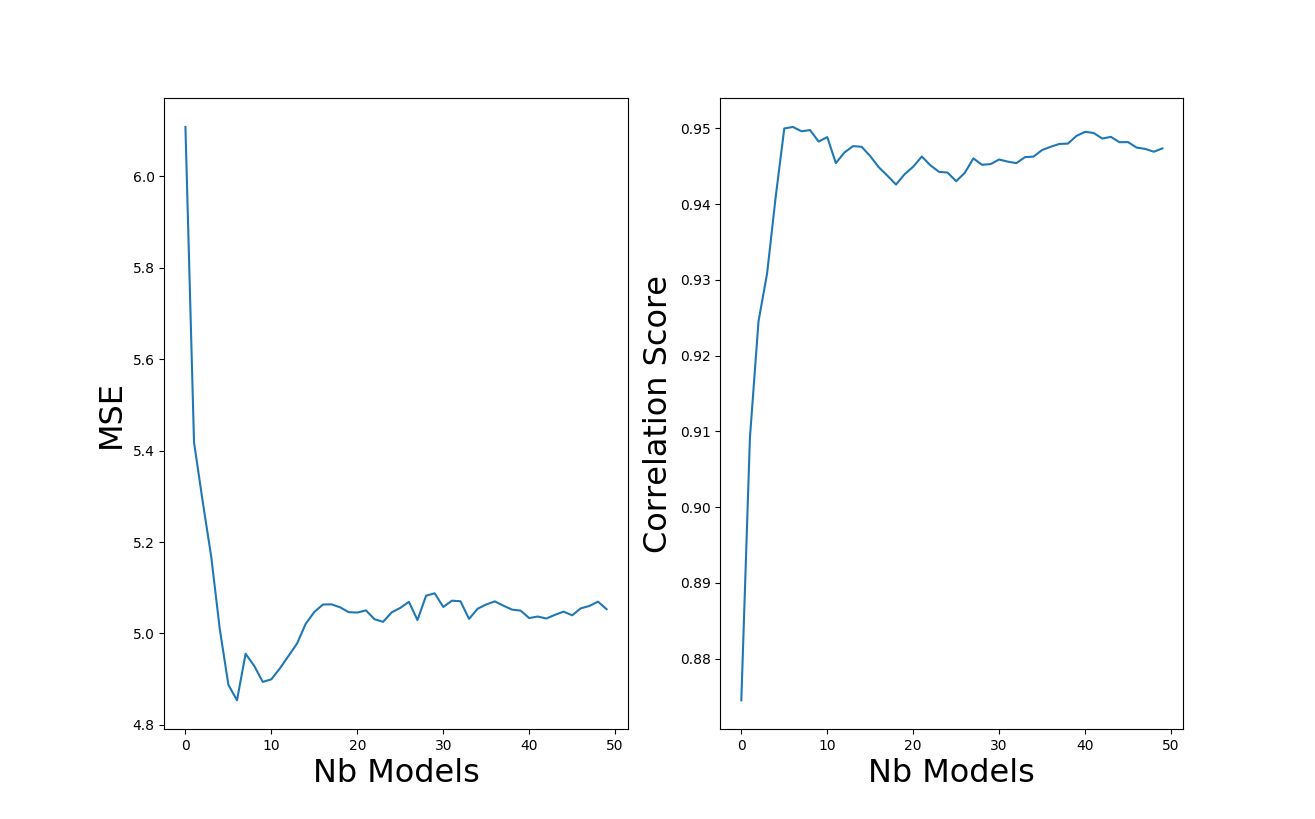}
    \caption{Evolution of MSE and Correlation according to the numbers of averaged models.}
    \label{fig:ev_mse}
\end{figure}

To study AGRA behaviour, it is interesting to show the evolution of both MSE and Pearson Correlation, according to the number of averaged models (Figure~\ref{fig:ev_mse}).
As stated before, gradients are model-dependant.
So, MSE, Pearson coefficient and thus the explanation of the network change a lot according to the model. 
More particularly, it can been seen in Figure~\ref{fig:ev_mse} that the two first training lead to bad results while the following ones, before the tenth, have a good explanation. Let's remember that the different model changes just by the initialization of the weights. They all have nearly the same regression scores but their gradients change strongly.
It is therefore impossible to define \textit{a priori} the models that lead to a good quality gradient.
So, in Figure~\ref{fig:ev_mse}, the $MSE$ is important at the beginning and then decreases before stabilizing. 
Averaging the gradients obtained by 20 or more models produces good explanation results, independent of learning.
The same reasoning can be applied to Pearson correlation coefficient.

\subsection{AGRA combined with other methods}
As stated before, it is possible to combine our approach with different state-of-the-art methods, such as GRAD$\times$Input, Smooth Grad and Integrated gradient, in order to improve both qualitative and quantitative results.

\begin{figure*}
\centering
  \includegraphics[width=.85\textwidth]{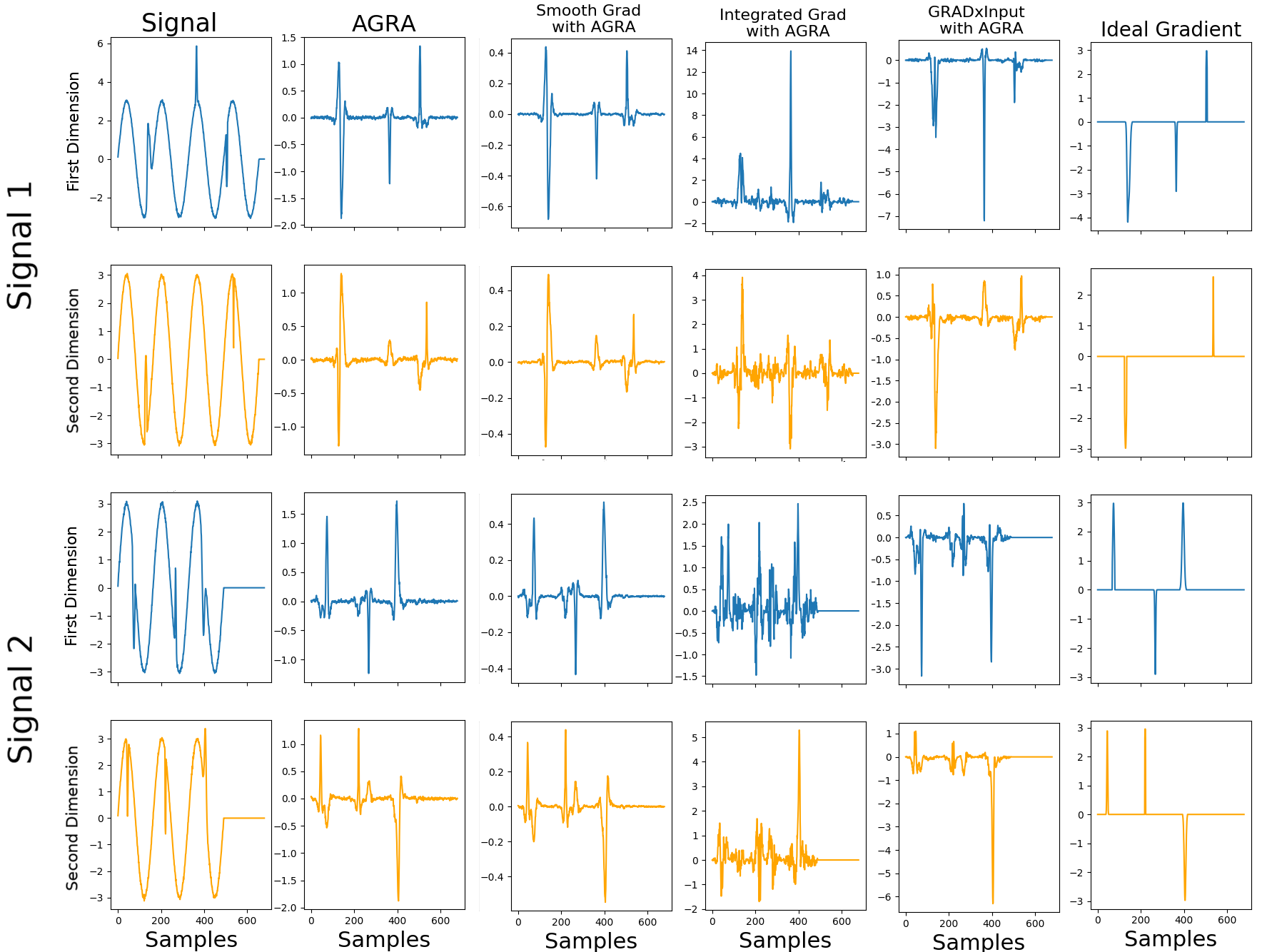}
  \caption{Results obtained with all the methods averaged over 50 model over the 2 same examples.}
  \label{fig:all_agra}
\end{figure*}

As shown in Figure~\ref{fig:all_agra}, using the average of $50$ models for all methods greatly improves their performances and especially denoises results of every methods.
Quantitative results  are all improved using AGRA as shown in Table~\ref{tab:all_agra}, for both Pearson correlation and MSE.  
This shows that even if this method is computationally intensive, obtained results are really improved compared with state-of-the-art.
\begin{table}[]
    \centering
    \begin{tabular}{c|c|c}
        \hline
         \bfseries Methods & \bfseries Pearson Correlation & \bfseries MSE\\
         \hline
         \thead{GRAD \cite{simonyan2013deep}} & 0.81 & 7.65 \\
         \thead{AGRA} & \bfseries 0.94 & \bfseries 5.06 \\
         \hline
         \thead{GRAD$\times$Input \cite{shrikumar2017learning}, \cite{bach2015pixel}} & 0.82 & NA \\
         \thead{GRAD$\times$Input with AGRA} & 0.89 &  NA \\
        \hline
         \thead{Smooth Grad \cite{smilkov2017smoothgrad}} & 0.79 & 7.85 \\
         \thead{Smooth Grad \\with AGRA} & 0.92 & 6.91  \\
        \hline
         Integrated Gradient \cite{sundararajan2017axiomatic} & 0.55 & NA \\
         \thead{Integrated Gradient \\with AGRA} & 0.82 & NA \\
         \hline
    \end{tabular}
    \vspace{.3cm}
    \caption{Pearson correlation coefficient and MSE for different methods combined with AGRA framework.}
    \label{tab:all_agra}
\end{table}

\section{Conclusion}
 In this paper a new approach to explain neural network decisions has been presented, with a specific experimental setup dedicated to neural network explanation.
 Indeed, the lack of ground truth for network explanation often only allows a qualitative comparison of different approaches.  The design of a synthesis device, devoted to this task, enables quantitative comparisons.
 
 In addition to this new database and experimental setup, a novel approach for network decision explanation has been proposed. 
 Indeed, by observing that the explanation strongly depends on the learning of the model, we proposed to carry out several trainings and then to average explanations provided by each of them. It has been shown that this technique improves both qualitative results - indeed explanations are less noisy - and quantitative results, with better scores for both Pearson correlation and MSE of reconstructed signals. 
 However the drawback of this method, is the high computation cost, since many models need to be trained. 
 
 In the future, we plan to extend this approach to models learned in classification to see if the same conclusions can be drawn.

\bibliographystyle{IEEEtran}
\bibliography{IEEEabrv}

\end{document}